\documentclass[sigconf, nonacm]{acmart}
\AtBeginDocument{%
  }

\setcopyright{acmlicensed}
\copyrightyear{2026}
\acmYear{2026}
\acmDOI{XXXXXXX.XXXXXXX}
\acmConference[Conference on Multimedia]{Make sure to enter the correct
  conference title from your rights confirmation email}{November 10–14,
  2026}{Rio de Janeiro, Brazil}
\acmISBN{978-1-4503-XXXX-X/2018/06}

\usepackage{colortbl}
\usepackage{pifont}

\begin{document}

\title{One CT Unified Model Training Framework to Rule All Scanning Protocols}


\author{Fengzhi Xu}
\affiliation{%
  \institution{Sichuan University}
  \city{Chengdu}
  \country{China}}

\author{Ziyuan Yang}
\affiliation{%
  \institution{Sichuan University}
  \city{Chengdu}
  \country{China}
}

\author{Zexin Lu}
\affiliation{%
 \institution{Sichuan University}
 \city{Chengdu}
 \country{China}}

\author{Yingyu Chen}
\affiliation{%
 \institution{Sichuan University}
 \city{Chengdu}
 \country{China}}

\author{Fenglei Fan}
\affiliation{%
  \institution{City University of Hong Kong}
  \city{Hong Kong}
  \country{China}}

\author{Hongming Shan}
\affiliation{%
  \institution{Fudan University}
  \city{Shanghai}
  \country{China}}

\author{Yi Zhang}
\affiliation{%
 \institution{Sichuan University}
 \city{Chengdu}
 \country{China}}

\renewcommand{\shortauthors}{Trovato et al.}

\begin{abstract}
Non-ideal measurement computed tomography (NICT), which lowers radiation at the cost of image quality, is expanding the clinical use of CT. Although unified models have shown promise in NICT enhancement, most methods require paired data, which is an impractical demand due to inevitable organ motion. Unsupervised approaches attempt to overcome this limitation, but their assumption of homogeneous noise neglects the variability of scanning protocols, leading to poor generalization and potential model collapse. We further observe that distinct scanning protocols, which correspond to different physical imaging processes, produce discrete sub-manifolds in the feature space, contradicting these assumptions and limiting their effectiveness. To address this, we propose an Uncertainty-Guided Manifold Smoothing (UMS) framework to bridge the gaps between sub-manifolds. A classifier in UMS identifies sub-manifolds and predicts uncertainty scores, which guide the generation of diverse samples across the entire manifold. By leveraging the classifier’s capability, UMS effectively fills the gaps between discrete sub-manifolds, and promotes a continuous and dense feature space. Due to the complexity of the global manifold, it's hard to directly model it. Therefore, we propose to dynamically incorporate the global- and sub-manifold-specific features. Specifically, we design a global- and sub-manifold-driven architecture guided by the classifier, which enables dynamic adaptation to subdomain variations. This dynamic mechanism improves the network’s capacity to capture both shared and domain-specific features, thereby improving reconstruction performance. Extensive experiments on public datasets are conducted to validate the effectiveness of our method across different generation paradigms.
\end{abstract}




\maketitle

\section{Introduction}
\label{sec:intro}
Computed tomography (CT) is a widely adopted non-invasive imaging modality that offers detailed anatomical information, but it exposes patients to radiation, which raises safety concerns~\cite{koetzier2023deep, ria2024optimization}. To alleviate this, various scanning protocols are used to reduce radiation dose, including low-dose CT~(LDCT), sparse-view CT~(SVCT), and limited-angle CT~(LACT)~\cite{yang2022learning}. However, the measured data acquired using these protocols are considered non-ideal measurement CT~(NICT), and the reconstructed images are often degraded by artifacts and noise, which significantly compromise their clinical applicability.

To improve image quality and ensure clinical utility, unified models for NICT enhancement have demonstrated remarkable performance~\citep{10253669, du2024dper, pai2025vision, wang2025misd}. However, most of these methods rely on the availability of paired datasets consisting of NICT and ideal measurement CT~(ICT) images for training~\citep{yang2018low, xia2023regformer}. This assumption is difficult to meet due to the physiological processes, such as respiration and cardiac motion, which cause organ displacement and make it nearly impossible to acquire paired data.

To address this issue, unsupervised image quality enhancement methods have been developed to eliminate the need for paired data~\citep{kwon2021cycle, kim2024unsupervised}. Many of these approaches are based on variants of CycleGAN~\cite{zhu2017unpaired}, designed to learn mappings between the low- and high-quality image domains. For example, GcGAN~\cite{fu2019geometry} uses a geometry-consistency constraint by inputting an image and its transformed counterpart to reduce the solution space. CUT~\cite{park2020contrastive} improves performance with a patch-based contrastive learning framework that maximizes mutual information between input and output patches. AttentionGAN~\cite{tang2021attentiongan} enhances the generation quality by incorporating an attention mechanism. More recently, UNSB~\cite{kim2024unpaired} addresses the limitations of diffusion models for unpaired mapping by reformulating the Schrödinger Bridge problem as a sequence of adversarial learning tasks. CSUD~\cite{dong2025channel} performs unsupervised image deraining by incorporating a channel consistency prior and a self-reconstruction strategy to mitigate the reliance on paired data and improve the robustness. However, most methods assume a homogeneous source-domain distribution, which rarely holds in practice, as different scanning protocols are applied according to patient conditions and clinical guidelines.

\begin{figure}[t]
  \centering
  \includegraphics[width=\linewidth]{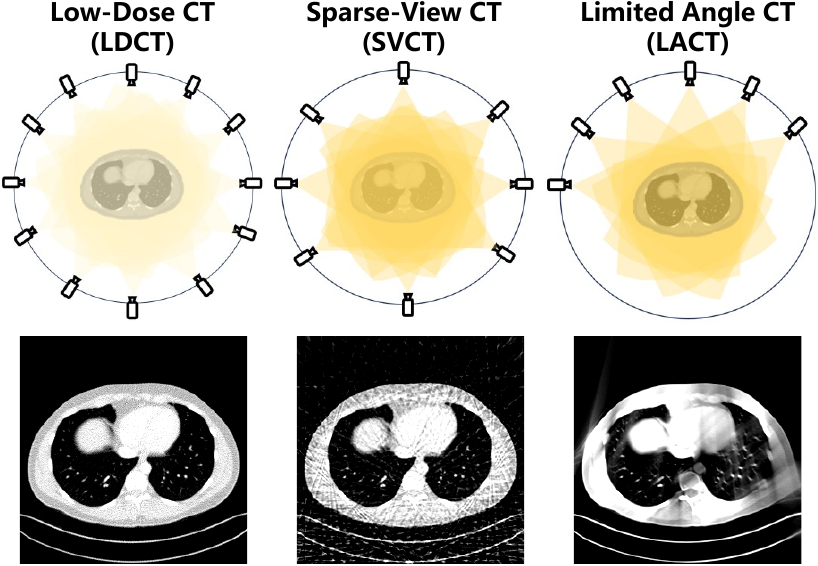}
  \caption{Visualization of reconstructed results under different scanning protocols.}
  \label{VisofNoise}
\end{figure}

Different scanning protocols correspond to distinct physical imaging processes, resulting in variations in the associated noise distributions~\citep{yang2025patient, du2024dper}. For instance, LDCT reduces radiation dose by lowering X-ray intensity. In contrast, both SVCT and LACT achieve dose reduction by decreasing the number of projections, but their acquisition patterns differ: SVCT preserves full angular coverage with reduced projection density, whereas LACT confines acquisition to a limited angular range. As a result, each scanning protocol exhibits a distinct noise distribution. To help readers better understand it, we illustrate the scanning protocols and their corresponding reconstructed images in Figure~\ref{VisofNoise}. 

Besides, patients differ in their anatomical composition, which leads to variations in the distribution of tissue-specific attenuation coefficients~\cite{shi2025zs4d}. Since CT essentially measures these attenuation properties. such anatomical differences further amplifies the heterogeneity observed in CT images. Hence, in this paper, we aim to answer the following question:

{\textit{``Can we develop an unsupervised, general training framework that models heterogeneous tissue distribution patterns across patients and scanning protocols, thereby enabling high-quality reconstruction under reduced scanning conditions?"}}

To achieve this, we train a unified model using data from multiple patients scanned under different protocols, rather than training a separate model for each protocol. This unified setting enables the model to leverage richer anatomical information while avoiding the additional training and storage costs associated with protocol-specific models. However, accomplishing this goal requires addressing the inherent heterogeneity introduced by the diverse noise characteristics across different scanning protocols.

To alleviate this issue, we model the feature representations of images acquired via different scanning protocols as points distributed on a complex manifold. Due to fundamentally different physical scanning processes, these points naturally cluster into discrete sub-manifolds corresponding to each scanning protocol. Formally, the manifold of a non-ideal data domain can be modeled as a union of these sub-manifolds. However, these sub-manifolds are often disconnected and discrete, which leads to discontinuities across the global feature space. Such discontinuities pose challenges for learning models, as features from different sub-manifolds exhibit abrupt transitions without smooth and continuous trajectories connecting them.

To address this issue, we propose an Uncertainty-Guided Manifold Smoothing~(UMS) framework for unified model training. Specifically, a classifier is first trained to identify the sub-manifold associated with each feature representation, and its predictions are used to guide the diffusion process for generating diverse samples across the entire manifold. By incorporating the classifier’s uncertainty estimates, particularly near sub-manifold boundaries, the framework effectively bridges the gaps between discrete sub-manifolds. This approach promotes a more continuous and dense feature space, which facilitates smoother transitions across sub-manifold boundaries.

Although the manifold smoothing process forms a more continuous global manifold, substantial differences among
sub-manifolds remain. This discrepancy poses challenges for a single model trained on the entire manifold to effectively capture the specific knowledge and nuances of each sub-manifold. As a result, relying solely on a global representation risks overlooking important sub-manifold-specific characteristics, which compromise the accuracy and robustness of reconstruction. To address this challenge, we design a global- and sub-manifold-driven architecture. This architecture is guided by the same classifier employed during the manifold smoothing stage, which controls the generation process. Importantly, in practice, modeling the complex relationships between scanning protocols, defined by various physical parameters, and the resulting noise distributions is challenging due to the complex physical imaging process. Instead, our method leverages the classifier’s predictions in the image domain to infer latent relationships among subdomains. Such guidance enables the model to dynamically adapt to sub-manifold-specific features while maintaining consistency with the global manifold, thus benefiting from shared global- and sub-manifold knowledge. The main contributions of this paper can be summarized as follows:

\begin{itemize}
    \item We introduce a novel uncertainty-guided training framework for unified models that effectively bridges the gaps between discrete sub-manifolds in the NICT feature space.

    \item We design a confidence-guided global- and sub-manifold-driven architecture that jointly models and balances the latent relationships between global- and sub-manifold feature representations.

    \item Extensive experiments on public datasets validate the effectiveness of the proposed method, and the results demonstrate its compatibility with various approaches to further enhance reconstruction performance.
\end{itemize}

\section{Methodology}
\label{sec:met}
\subsection{Overview}
\begin{figure*}[!t]
\centering
\includegraphics[width=\textwidth]{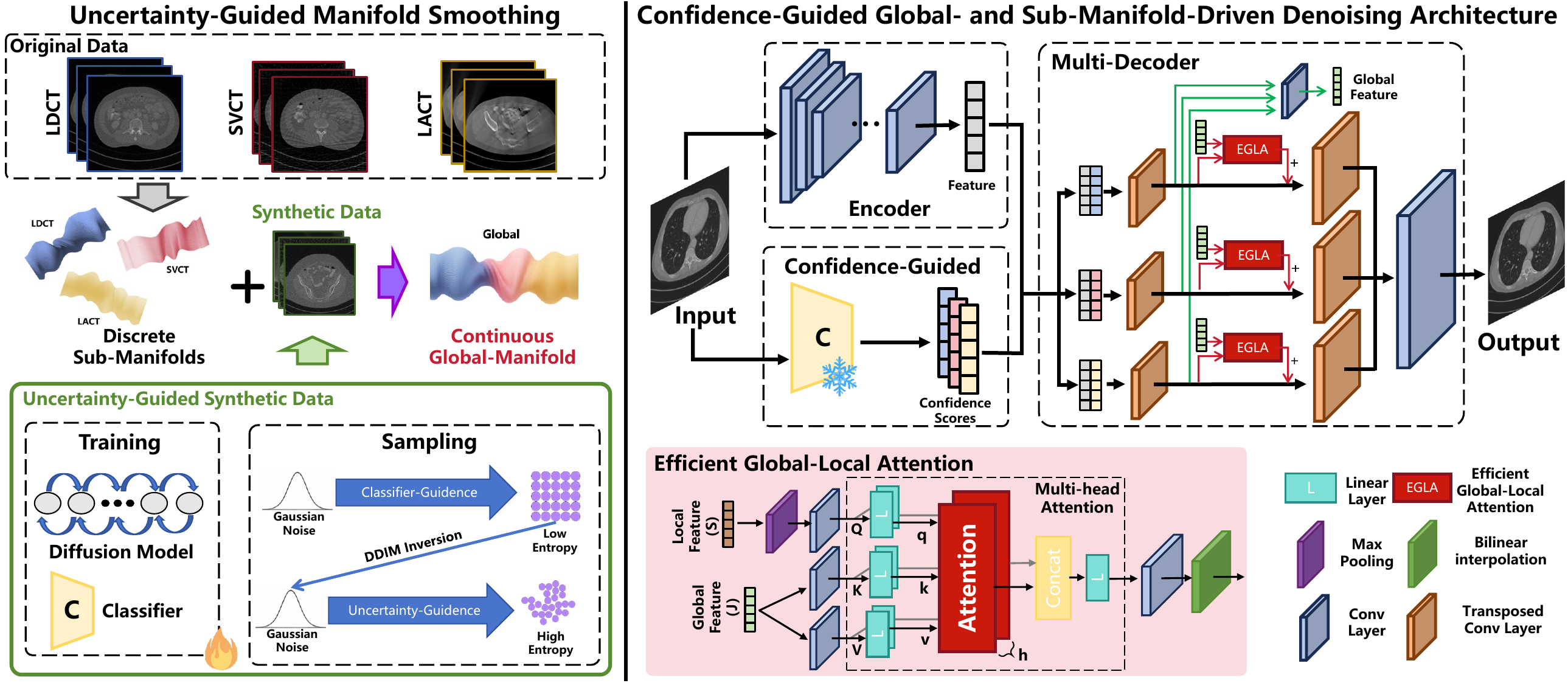} 
\caption{The overview of the proposed framework.}
\label{framework}
\end{figure*}

The overview of our proposed training framework can be found in Figure~\ref{framework}. Our framework consists of two stages. The first stage trains a classifier using data from different sub-manifolds. The uncertainty scores produced by this classifier guide the generation of new samples to bridge the gaps between sub-manifolds. Although the completed data form a global manifold, modeling it globally tends to overlook local features. To address this, we propose a confidence-guided global- and sub-manifold architecture in the second stage. This structure leverages the classifier’s predictions to infer implicit relationships between sub-manifolds, which enable the network to dynamically adapt to subdomain-specific features. Additionally, a global-local attention module is designed to integrate global- and sub-manifold information effectively. 

\subsection{Uncertainty-Guided Manifold Smoothing}
As mentioned earlier, the proposed smoothing step aims to mitigate the domain gap and enhance the smoothness of sub-manifolds. This module operates in two stages: training and sampling. We now elaborate on its implementation in each stage.

\noindent \textbf{{Training Stage.}}
The diffusion model is trained by adding Gaussian noise $\epsilon$ to the data $x$ in a forward process, followed by learning to reverse this corruption using a time-conditioned U-Net architecture, which serves as the noise predictor conditioned on label information $y$, with its output denoted as $\epsilon_{\theta}$. Given a variance schedule ${\beta_t}$, we can sample the noisy image $x_t$ at any timestep $t$ using the reparameterization trick, where $x_t \sim \mathcal{N}(x_t; \sqrt{\overline{\alpha}_t}x_0, (1 - \overline{\alpha}_t) \mathbf{I})$, with $\alpha_t = 1 - \beta_t$ and $\overline{\alpha}_t = \prod_{i=1}^{t} \alpha_i$. By incorporating label information during training, the model learns to generate class-related samples~\citep{ho2022classifier}.

The diffusion training loss function is as follows:
\begin{equation}
\begin{split}
\mathcal{L}_{\text{diff}} &= \sum_{t}\mathbb{E}_{x_0,\epsilon}[W||\epsilon - \epsilon_{\theta}(x_t, y, t)||^2] + \zeta\mathcal{L}_{vlb},
\end{split}
\end{equation}
where $W$ is a time-step-dependent weight that facilitates the learning of visual information, $\zeta$ is a hyperparameter set to $0.001$ and $L_{vlb}$ introduces a learnable variance into the variational lower bound (VLB) loss to enhance the generative performance of the trained model~\citep{nichol2021improved}.

In addition, implementing guidance sampling requires pre-training a classifier $p_{\phi}(y|x_t, t)$, which incorporates time information $t$ as input to represent the degree of noise corruption in the noisy image. The classifier is based on the encoder part of a U-Net architecture, which incorporates attention blocks and timestep embeddings. The corresponding training loss function associated with this process is given by:
\begin{equation}
    \mathcal{L}_{\text{pre-classifier}} = \sum_t \mathbb{E}_{x_0, \epsilon}[||y-p_{\phi}(x_t,t)||_c],
\end{equation}
where $||.||_c$ represents the cross-entropy loss function.

\noindent \textbf{{Classifier-Guided Sampling and DDIM Inversion.}}
To ensure the generated images belong to the specified category, the pretrained classifier guides the generation of high-confidence, class-consistent samples~\citep{dhariwal2021diffusion}, after which reverse Denoising Diffusion Implicit Model~(DDIM) is applied to derive the corresponding noise data. The noise prediction process is formulated as follows:
\begin{equation}
    \hat{\epsilon} = \epsilon_{\theta}(x_t, y, t) - \sqrt{1-\bar{\alpha}_t}\cdot \nabla_{x_t}\log p_{\phi}(x_t, t).
\end{equation}
Based on the gradient-guided noise prediction, a new sampling process is derived as:
\begin{equation}
    x_{t-1} = \sqrt{\bar{\alpha}_{t-1}}\left(\frac{x_t - \sqrt{1-\bar{\alpha}_t } \hat{\epsilon}}{\sqrt{\bar{\alpha}_t}}\right) + \sqrt{1-\bar{\alpha}_{t-1}}\hat{\epsilon}.
\end{equation}

To proceed to the next stage of generation, we generate new noisy data by reversing the deterministic generative process of DDIM:
\begin{equation}
\begin{split}
    x_{t+1} &= \sqrt{\bar{\alpha}_{t+1}}\left( \frac{x_t - \sqrt{1-\bar{\alpha}_t} \epsilon_{\theta}(x_t,y,t)}{\sqrt{\bar{\alpha}_t}} \right) \\
    &\quad + \sqrt{1-\bar{\alpha}_{t+1}}\epsilon_{\theta}(x_t,y,t).
\end{split}
\end{equation}

\noindent \textbf{{Uncertainty-Guided Sampling.}}
To improve the diversity of generated samples and promote their distribution near sub-manifold boundaries to ensure a smooth transition across sub-manifolds. Building upon the classifier guidance, we extend it to develop an uncertainty-guided sampling strategy.

From the perspective of stochastic differential equation~(SDE), the classifier is more intuitive and naturally supports extending classifier guidance. 
When the sampling variance is zero, the following DDIM sampling process is obtained:
\begin{equation}
\mathrm{d}x = \left(f_t(x_t) - \frac{1}{2}g^2_t \nabla_{x_t}\log p_{\theta}(x_t)\right)\mathrm{d}t,
\end{equation}
where $\nabla_{x_t}\log p_{\theta}(x_t)$ is score function, $f_t(\cdot)$ is the drift coefficient of $x_t$, $g_t$ is the diffusion coefficient of $x_t$. Similar to class label or image/text embeddings, we introduces an uncertainty score $u$ as a condition.
Then, the conditional ODE of uncertainty guidance in DDIM is
\begin{equation}
\mathrm{d}x = \left( f_t(x_t)-\frac{1}{2}g_t^2\nabla_{x_t}\log p_{\theta}(x_t|u) \right) \mathrm{d}t,
\end{equation}
wherein, exploiting the Bayesian formula in the score function and selecting terms about $x_t$, we obtain:
\begin{align}
\nabla_{x_t} \log p_{\theta}(x_t \mid u)
    = \nabla_{x_t} \log \left( \frac{p_{\theta}(x_t)p(u \mid x_t)}{p(u)} \right) \nonumber \\
    = \nabla_{x_t} \log p_{\theta}(x_t) + \nabla_{x_t} \log p(u \mid x_t).
\end{align}
Next, we leverage the connection between diffusion models and score matching, and the score function can be expressed as:
\begin{equation}
\nabla_{x_t}\log p_{\theta}(x_t) = -\frac{1}{\sqrt{1-\bar{\alpha}_t}}\epsilon_{\theta}(x_t,y,t).
\end{equation}

Moreover, to obtain sampling results $x_0$ with higher uncertainty value, following \citep{luo2024measurement} we set $p(u|x_t) \propto e^{\gamma \cdot \mathcal{U}(x_t) }$, where $\gamma$ is a hyperparameter that controls the strength of uncertainty guidance and $\mathcal{U}(\cdot)$, derived from the pre-trained classifier $p_{\phi}(y|x_t, t)$, represents entropy used for uncertainty-guided and is defined as follows:
\begin{equation}
    \mathcal{U}(x_t) = - \sum\limits_{i}p_{\phi}(y_i|x_t,t) \log p_{\phi}(y_i|x_t,t).
\end{equation}
This leads to $p_{\theta}(x_t|u) \propto e^{\gamma \cdot \mathcal{U}(x_t)}p_{\theta}(x_t)$, thereby encouraging the model to generate samples with higher uncertainty values. Furthermore,
\begin{align}
&\nabla_{x_t} \log p_{\theta}(x_t\mid u)\!=\!-\frac{1}{\sqrt{1\!-\!\bar{\alpha}_t}} \epsilon_{\theta}(x_t,y,t) 
       \!+\!\nabla_{x_t} \gamma \cdot \mathcal{U}(x_t) \nonumber \\
    &=\!-\frac{1}{\sqrt{1\!-\!\bar{\alpha}_t}} \Bigl( \epsilon_{\theta}(x_t,y,t) 
       \!-\!\sqrt{1\!-\!\bar{\alpha}_t}\!\cdot\!\nabla_{x_t} \gamma\!\cdot\!\mathcal{U}(x_t)\!\Bigr),
\end{align}

Analogous to classifier guidance, we introduce uncertainty-guided sampling by replacing $\epsilon_{\theta}(x_t, y, t)$ in each sampling step with
\begin{equation}
\hat{\epsilon}_{\theta}(x_t, y, t) = \epsilon_{\theta}(x_t, y, t) - \sqrt{1 - \overline{\alpha}_t} \cdot \nabla_{x_t} \gamma \cdot \mathcal{U}(x_t).
\end{equation}

Consequently, the new sampling process tends to generate $x_t$ with larger $\mathcal{U}(x_t)$ values, which ultimately results in a $x_0$ with a larger $\mathcal{U}(x_0)$, leading to more uncertain information.

\subsection{Network Architecture}
Although the manifold has been smoothed by the first step, modeling the global manifold with a single representation remains challenging due to inherent differences among the sub-manifolds. To address this issue, we propose a confidence-guided denoising architecture that leverages both global and sub-manifold information.
Our framework consists of two generators and two discriminators. The generator that maps from the high-quality to the low-quality domain, along with the two discriminators, follows the design of previous works. For the generator that maps from the low-quality to the high-quality domain, we propose a novel architecture that models both global- and sub-manifolds. In other words, our method can be seamlessly integrated into existing unsupervised frameworks by simply modifying the generator architecture.

Existing generators for low- to high-quality mapping typically rely on feature processing based on a single, simple manifold, which limits their capacity to capture the complexity of manifolds, such as the smoothed one in our study. To enable the generator to adapt to complex manifolds characterized by intrinsic differences, we rethink the mapping process from the low-quality to the high-quality domain. These differences primarily arise from noise introduced by different physical scanning procedures. To address this, we model the image in the feature domain as a superposition of multiple sub-manifolds and employ a classifier to obtain confidence scores that implicitly capture their interrelationships. Furthermore, we design multiple decoders, each tailored to the feature processing of different sub-manifolds, and use the confidence scores to guide this process, thereby achieving adaptive feature representation and processing.

Specifically, the generator consists of an encoder and multiple decoders. The encoder extracts features from the input image. To reduce training cost, we reuse the classifier trained within the UMS framework. The classifier assigns a soft label to each image, and the resulting confidence scores are concatenated with the image features as input to the decoder module. The input to each decoder, denoted as $I_d$, is defined as:
\begin{equation}
    I_d = Concat(f, c_d),
\end{equation}
where $f$ denotes the image feature extracted by the encoder, and $c_d$ represents the confidence score for a sub-manifold, reshaped to match the dimensions of $f$. The multiple decoders produce outputs $o_1, o_2, o_3$ corresponding to different sub-manifolds. Finally, the outputs are fused to generate the final high-quality image $O$ as follows:
\begin{equation}
    O = Conv(Avg(o_1, o_2, o_3)).
\end{equation}

To enable more effective feature aggregation across multiple decoder outputs, we introduce a global-local attention module, which allows the sub-manifold-driven decoders to incorporate global feature information during processing. This helps mitigate excessive bias and promotes more reliable aggregation in subsequent stages.

The global-local attention module comprises two components: global feature representation and efficient global-local attention~(EGLA). The global feature representation processes the feature maps from the first transposed convolution layer of all decoders to generate a global-manifold representation $J\in \mathbb{R}^{B\times C \times H \times W}$. Each decoder integrates an EGLA module between two transposed convolution layers. The EGLA module receives the global feature $J$ and the intermediate feature map $S\in \mathbb{R}^{B\times C \times 2H \times 2W}$, obtained after the first transposed convolution layer.

The structure of the EGLA module is illustrated in Figure~\ref{framework}. The core attention mechanism within EGLA takes three inputs: query $q\in \mathbb{R}^{B\times N \times d_k}$, key $k\in \mathbb{R}^{B\times N \times d_k}$, and value $v\in \mathbb{R}^{B\times N \times d_v}$. Here, $B$ represents the batch size, and $N=H\times W$ is the total number of spatial locations. The attention is calculated as follows:
\begin{equation}
    \text{Attention}(q, k ,v) = \text{softmax}(\frac{qk^{T}}{\sqrt{d_k}})v.
\end{equation}

However, a single attention mechanism captures relationships within one pattern, limiting its ability to model diverse associations. Multi-head attention addresses this by employing multiple parallel heads to capture interactions across different subspaces, thereby enhancing representation capability. We thus adopt the multi-head attention mechanism for feature fusion. The multi-head attention module receives three inputs: query $Q\in \mathbb{R}^{B\times N \times E}$, key $K\in \mathbb{R}^{B\times N \times E}$, and value $V\in \mathbb{R}^{B\times N \times E}$, where $E$ denotes the embedding dimension. 

To reduce computational cost, the feature map $S$ produced by the transposed convolution is downsampled via a max pooling layer, followed by a convolutional layer and flattening operation to generate the multi-head attention query $Q$. Meanwhile, the global feature $J$ is processed by two separate convolutional layers and then flattened to generate the key $K$ and value $V$, respectively. Additionally, the resulting attention feature map is upsampled via interpolation to match the dimensions of the convolutional feature map from the first transposed convolution, enabling a residual connection between them.

Our confidence-guided global- and sub-
manifold-driven denoising network introduces architectural innovations that can flexibly integrate with various loss functions. In this study, we use a conventional cycle-consistency-based loss function to facilitate network training, formulated below:
\begin{equation}
    \mathcal{L}(G,D) = \mathcal{L}_{\text{GAN}}(G,D) + \lambda_1\mathcal{L}_{\text{cyc}}(G)+\lambda_2\mathcal{L}_{\text{ide}}(G),
\end{equation}
where $G$ and $D$ denote the generator and discriminator, respectively. $L_{GAN}$ represents the adversarial loss used to train the generator and discriminator in an adversarial manner. $L_{\text{cyc}}$ denotes the cycle consistency loss, which enforces cross-domain reconstruction consistency. $L_{\text{ide}}$ corresponds to the identity loss, which preserves within-domain reconstruction consistency.

\begin{figure}[!t]
    \centering
    \includegraphics[width=\columnwidth]{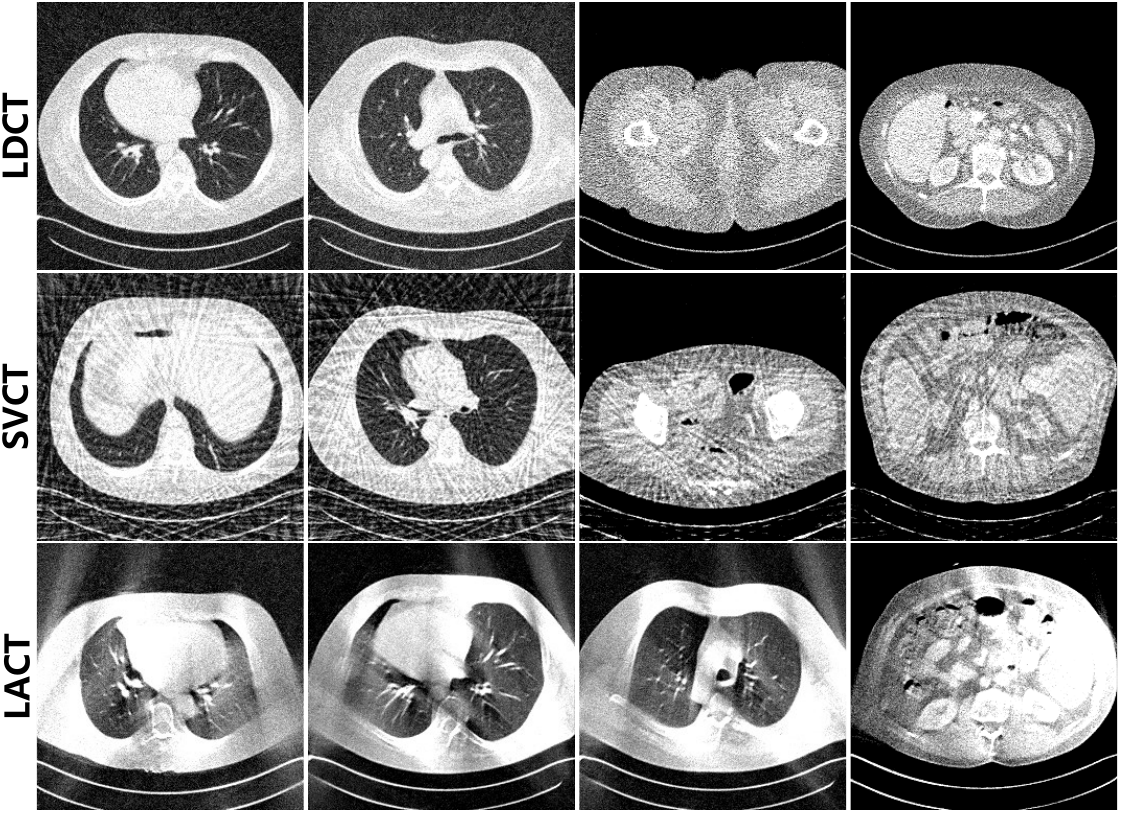}
    \caption{Generation examples of the uncertainty-guided diffusion model.}
    \label{generate_results}
\end{figure}

\begin{table}[!t]
\centering
\footnotesize
\caption{Scanning protocols used to simulate different NICT images.}
\begin{tabular}{ccccc}
	\toprule
         & & \multicolumn{3}{c}{Geometrical Parameters $\beta$} \\
        \cmidrule(l){3-5}
	Scanning Protocol & Photon Count $\alpha$ & View Number & Angle Range\\
	\midrule
	LDCT & $1.25e^{4}$ & 512 & $[0^\circ, 360^\circ]$\\
	SVCT & $1.25e^{8}$ & 60 & $[0^\circ, 360^\circ]$\\
	LACT & $1.25e^{8}$ & 512 & $[0^\circ, 125^\circ]$\\
	\bottomrule
\end{tabular}
\label{scangeo}
\end{table}

\begin{figure*}[!t]
    \centering
    \includegraphics[width=\textwidth]{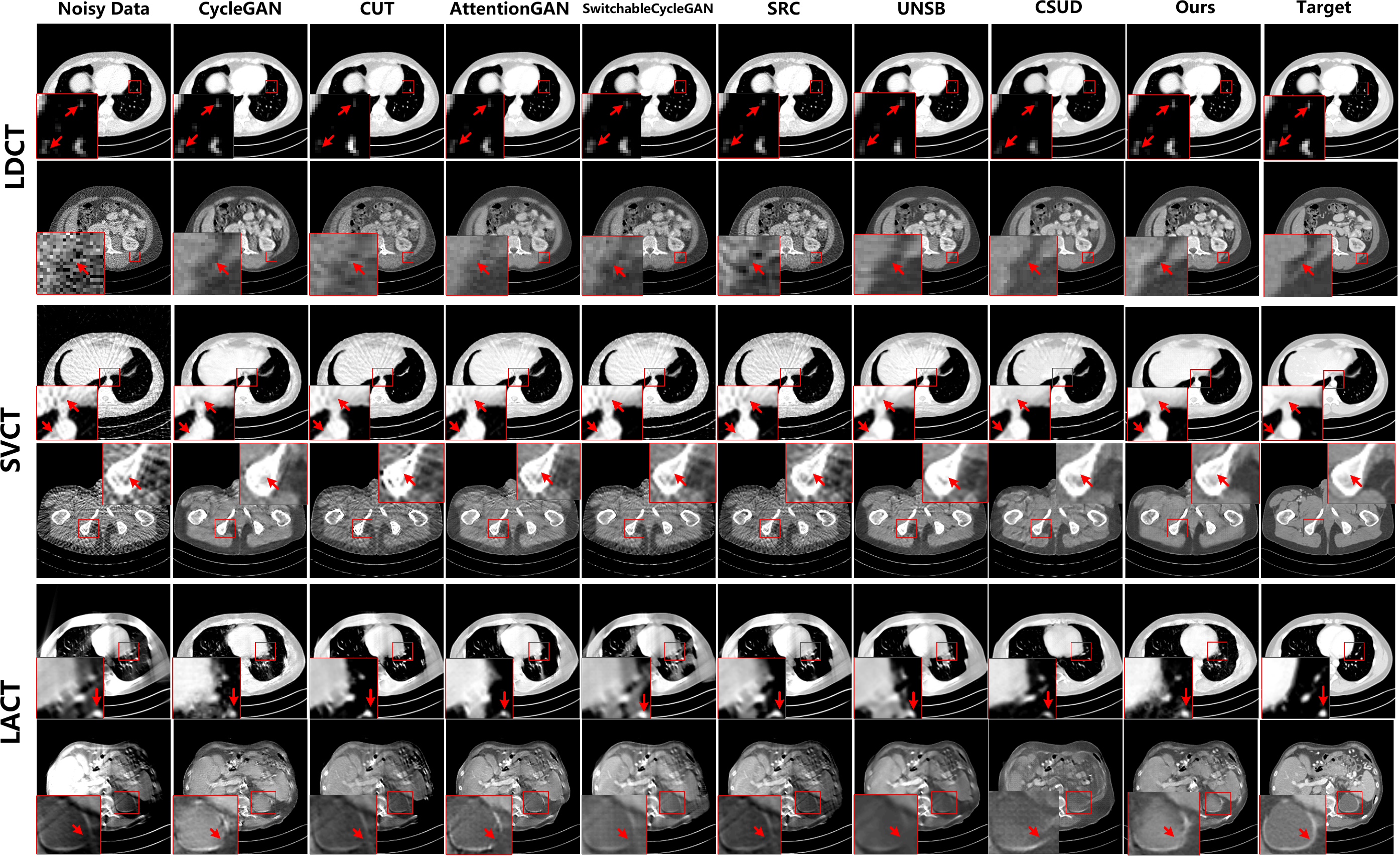}
    \caption{Reconstruction results of different algorithms under varying sampling strategies. The display window for rows 1, 3, and 5 is [-805, 145] HU, while for other rows, it is [-265, 285] HU.}
    \label{results_all}
\end{figure*}

\section{Experiments}
\subsection{Experimental Setup}
\noindent \textbf{{Datasets.}}
The publicly available “2016 NIH-AAPM-Mayo Clinic Low-Dose CT Grand Challenge” dataset \citep{mccollough2016tu} is used to evaluate the effectiveness of the proposed method. To reduce sample redundancy and improve training efficiency, a total of 193 ICT images were selected for training and 46 for testing.
Ultimately, our training dataset consists of 579 simulated and UMS-synthesized NICT images as the source domain, and 193 ICT images as the target domain.

\noindent \textbf{{Data Simulation.}}
Low-quality CT images are generated by undersampling the projection data in the projection domain during simulation. Different scanning protocols were employed to generate downsampled sinogram data, and then FBP was used to reconstruct LDCT, SVCT, and LACT images. The detailed parameters for each scanning protocol are summarized in Table~\ref{scangeo}. 

\begin{table*}[!t]
  \centering
  \caption{Quantitative results of PSNR(dB) $\uparrow$, SSIM(\%) $\uparrow$, LPIPS $\downarrow$ and NoiseSD $\downarrow$ for different algorithms under varying sampling strategies}
  \resizebox{\textwidth}{!}{
  \begin{tabular}{lcccccccccccc|cccc}
    \toprule
    &\multicolumn{4}{c}{LDCT} & \multicolumn{4}{c}{SVCT} & \multicolumn{4}{c}{LACT} & \multicolumn{4}{c}{Average}\\
    \cmidrule(r){2-5} \cmidrule(r){6-9} \cmidrule(r){10-13} \cmidrule(r){14-17}
    Algorithm & PSNR & SSIM & LPIPS & NoiseSD & PSNR & SSIM & LPIPS & NoiseSD & PSNR & SSIM & LPIPS & NoiseSD & PSNR & SSIM & LPIPS & NoiseSD\\
    \midrule
    CycleGAN & 38.25 & 94.57 & 0.0306 & 0.0120 & 36.01 & 91.44 & 0.0444 & 0.0157 & 31.97 & 90.93 & 0.0522 & 0.0253 & 35.41 & 92.31 & 0.0424 & 0.0176\\
    CUT & 36.21 & 90.84 & 0.0654 & 0.0151 & 33.37 & 81.92 & 0.1196 & 0.0211 & 29.42 & 83.48 & 0.0766 & 0.0344 & 33.00 & 85.41 & 0.0543 & 0.0235\\
    AttentionGAN & 39.59 & \textbf{95.83} & 0.0341 & 0.0108 & 36.70 & 90.91 & 0.0647 & 0.0148 & 31.01 & 91.06 & 0.0480 & 0.0258 & 35.76 & 92.60 & 0.0489 & 0.0171\\
    SwitchableCycleGAN & 37.26 & 92.92 & 0.0493 & 0.0129 & 33.38 & 84.39 & 0.1057 & 0.0206 & 27.40 & 85.45 & 0.1000 & 0.0414 & 32.68 & 87.58 & 0.0850 & 0.0249\\
    SRC & 35.52 & 88.16 & 0.0899 & 0.0164 & 32.07 & 79.05 & 0.1556 & 0.0244 & 29.09 & 83.92 & 0.0696 & 0.0351 & 32.22 & 83.71 & 0.1050 & 0.0253\\
    UNSB & 36.82 & 94.81 & 0.0405 & 0.0139 & 34.81 & 90.93 & 0.0582 & 0.0177 & 29.29 & 86.77 & 0.0805 & 0.0343 & 33.64 & 90.83 & 0.0597 & 0.0219\\
    CSUD & 36.08 & 93.22 & 0.0454 & 0.0155 & 33.71 & 89.18 & 0.0747 & 0.0206 & 29.74 & 88.04 & 0.0749 & 0.0330 & 33.17 & 90.14 & 0.0650 & 0.0230\\ \hline
    \rowcolor{gray!20} Ours & \textbf{39.66} & 95.18 & \textbf{0.0234} & \textbf{0.0102} & \textbf{36.89} & \textbf{92.34} & \textbf{0.0407} & \textbf{0.0142} & \textbf{33.79} & \textbf{92.06} & \textbf{0.0375} & \textbf{0.0207} & \textbf{36.78} & \textbf{93.19} & \textbf{0.0338} & \textbf{0.0150}\\
    \bottomrule
  \end{tabular}
  }
  \label{reconstruction results}
\end{table*}

\noindent \textbf{{Data Generation.}}
To smooth the manifold, the diffusion model and the classifier are trained separately using simulated low-quality data. Then, we apply classifier guidance with a scale of 10, followed by DDIM inversion to obtain the corresponding noise representation. Subsequently, uncertainty guidance is applied to the noise representation, with a guidance scale of 3. For each sub-manifold, 100 images are generated, and representative samples are presented in Figure~\ref{generate_results}.

\noindent \textbf{{Baselines.}}
The proposed method is compared against several baseline methods across two generation paradigms: GAN-based methods, including CycleGAN~\citep{zhu2017unpaired}, GcGAN~\citep{fu2019geometry}, CUT~\citep{park2020contrastive}, AttentionGAN~\citep{tang2021attentiongan}, SwitchableCycleGAN~\citep{yang2021continuous}, SRC~\citep{jung2022exploring}, and CSUD~\citep{dong2025channel}; and Schrödinger Bridge-based methods, including UNSB~\citep{kim2024unpaired}.

\noindent \textbf{{Implementation Details.}}
Following the settings in CycleGAN~\citep{zhu2017unpaired}, we set $\lambda_1=10$ and $\lambda_2=5$, and employed the Adam optimizer with $\beta_1 = 0.5$ and $\beta_2 = 0.999$ to train the proposed network. For the comparison methods, we used the released source codes 
for training and testing. The number of training epochs was set to 200. \footnote{The code will be made publicly available for reproducibility.}

\noindent \textbf{{Evaluation Metrics.}}
Consistent with previous works, Peak Signal-to-Noise Ratio (PSNR) and Structural Similarity (SSIM) are selected to evaluate the performance of different methods. For both metrics, higher values indicate better perceptual and structural image quality. Moreover, we employ Learned Perceptual Image Patch Similarity~(LPIPS) to assess high-level perceptual similarity and Noise Standard Deviation~(NoiseSD) to objectively quantify image noise.

\subsection{Comparison with Other Methods}
Figure~\ref{results_all} illustrates the qualitative comparisons across different methods and sampling conditions. While other methods achieve satisfactory performance under specific non-ideal sampling strategies, their generalization to other settings is limited, primarily due to the difficulty of modeling discrete sub-manifolds within a single unified model. In contrast, the proposed method maintains consistently high performance across all non-ideal sampling strategies. 

Quantitative results in Table~\ref{reconstruction results} further support the superiority of the proposed method. Across low-dose, sparse-view, and limited-angle scenarios, the proposed method consistently achieves the best reconstruction performance. The most significant improvement is observed in the limited-angle scenario, where the proposed method yields PSNR gains exceeding 1.8 dB compared to competing approaches. Additionally, it achieves notable improvements in average reconstruction quality across all three sub-manifolds.

\subsection{Generalization Evaluation}
\begin{table*}[!t]
  \centering
  \small
  \caption{Reconstruction results of different algorithms for domain generalization}
  \begin{tabular}{lcccccc|cccccc}
    \toprule
    &\multicolumn{4}{c}{Known Domain} & \multicolumn{2}{c}{Unkown Domain} &\multicolumn{4}{|c}{Known Domain} & \multicolumn{2}{c}{Unkown Domain}\\
    \cmidrule(r){2-5} \cmidrule(r){6-7} \cmidrule(r){8-11} \cmidrule(r){12-13}
     &\multicolumn{2}{c}{SVCT} & \multicolumn{2}{c}{LACT} & \multicolumn{2}{c}{LDCT} &
    \multicolumn{2}{|c}{LDCT} & \multicolumn{2}{c}{LACT} & \multicolumn{2}{c}{SVCT}\\
    \cmidrule(r){2-3} \cmidrule(r){4-5} \cmidrule(r){6-7} \cmidrule(r){8-9} \cmidrule(r){10-11} \cmidrule(r){12-13}
    Algorithm & PSNR & SSIM & PSNR & SSIM & PSNR & SSIM & PSNR & SSIM & PSNR & SSIM & PSNR & SSIM\\
    \midrule
    CycleGAN & 35.11 & 89.10 & 31.57 & 90.11 & 38.65 & 94.44 & 36.76 & 90.93 & 31.04 & 89.63 & 31.71 & 78.15 \\
    CUT & 32.25 & 77.11 & 28.69 & 81.03 & 36.48 & 91.75 & 35.92 & 87.26 & 28.21 & 81.61 & 31.18 & 73.33 \\
    SRC & 30.95 & 65.84 & 28.60 & 82.20 & 34.93 & 82.65 & 34.93 & 82.42 & 28.59 & 81.91 & 30.36 & 70.77 \\
    UNSB & 33.11 & 83.33 & 28.51 & 83.38 & 36.57 & 94.10 & 35.64 & 91.87 & 27.69 & 82.40 & 32.03 & 84.31 \\
    CSUD & 31.18 & 86.38 & 28.24 & 85.12 & 31.59 & 72.50 & 35.27 & 91.45 & 30.73 & 90.04 & 32.17 & 86.30\\
    \hline
    \rowcolor{gray!20} Ours & \textbf{35.88} & \textbf{91.55} & \textbf{33.46} & \textbf{91.32} & \textbf{39.00} & \textbf{94.79} & \textbf{39.53} & \textbf{95.39} & \textbf{33.15} & \textbf{91.35} & \textbf{33.46} & \textbf{86.06} \\
    \bottomrule
  \end{tabular}
  \label{results_domainGen}
\end{table*}
\noindent {\textbf{Domain Generalization Experiment.}} Due to the uncertainty-guided data generation and the new architecture, the proposed method can adapt to a broader range of data distributions, thereby enhancing its generalization ability. To evaluate the generalization performance on unknown domains, the model is trained on known domains and evaluated on both seen and unseen data distributions. The corresponding NICT image reconstruction results are summarized in Table \ref{results_domainGen}. As shown in Table~\ref{results_domainGen}, the proposed method consistently outperforms competing approaches on both known and unseen domains, which demonstrates strong generalization capability even when tested on previously unseen data.

\begin{figure}[!t]
    \centering
    \includegraphics[width=\columnwidth]{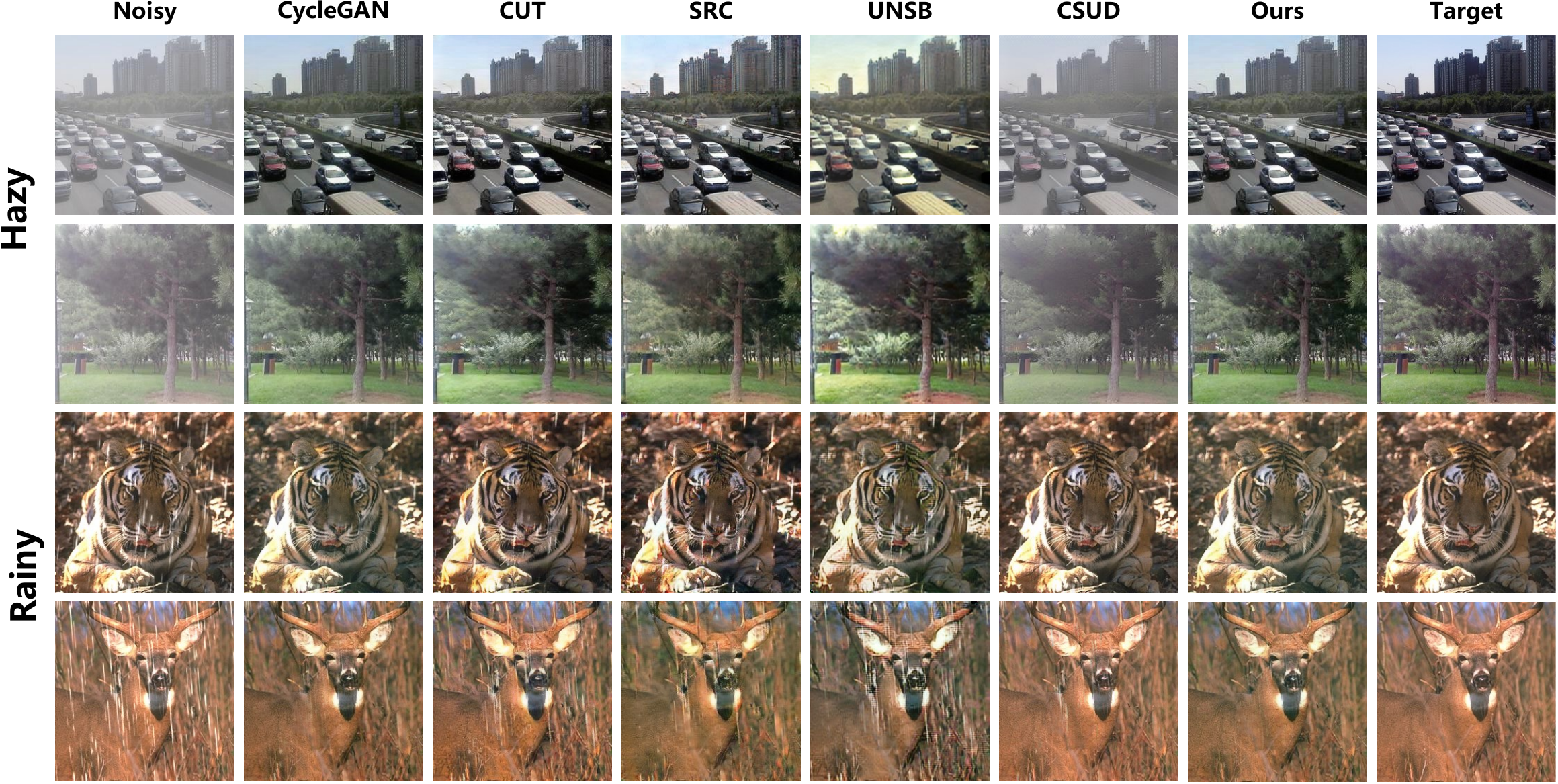}
    \caption{Reconstruction results of different algorithms on natural image dataset.}
    \label{results_natural}
\end{figure}

\noindent {\textbf{Natural Image Experiment.}} Moreover, to demonstrate the broader applicability of the proposed method beyond medical imaging, we evaluate its image reconstruction capability on natural image datasets. Specifically, we consider two natural image reconstruction tasks, image dehazing and image deraining tasks.

We construct the low-quality domain using 1,800 lightly rainy images from the JRDR dataset~\cite{jrdr} on Kaggle and 500 outdoor hazy images from the SOTS dataset~\citep{li2019benchmarking}, with their corresponding clean counterparts forming the high-quality domain. To balance the data distribution and validate our method, 1,000 additional hazy images are synthesized using UMS and added to the low-quality domain.

\begin{table}[!t]
  \centering
  \small
  \caption{Quantitative results for different algorithms on natural image dataset}
        \begin{tabular}{lllllll}
            \toprule
            &\multicolumn{2}{c}{Hazy} & \multicolumn{2}{c}{Rainy} & 
            \multicolumn{2}{c}{Average}\\
            \cmidrule(r){2-3} \cmidrule(r){4-5} \cmidrule(r){6-7} Algorithm & PSNR & SSIM & PSNR & SSIM & PSNR & SSIM\\
            \midrule
            CycleGAN & 29.18 & 95.31 & 25.31 & 88.97 & 27.24 & 92.14 \\
            CUT & 24.02 & 87.44 & 22.99 & 79.09 & 23.50 & 83.26 \\
            SRC & 22.80 & 88.07 & 20.32 & 73.55 & 21.56 & 80.81 \\
            UNSB & 22.70 & 83.33 & 21.31 & 72.65 & 22.00 & 77.99 \\
            CSUD & 23.13 & 82.11 & \textbf{30.30} & \textbf{92.81} & 26.71 & 87.46\\
            \hline
            \rowcolor{gray!20} Ours & \textbf{29.28} & \textbf{95.85} & 25.84 & 90.88 & \textbf{27.56} & \textbf{93.36} \\
            \bottomrule
        \end{tabular}
        \vspace{-15pt}
  \label{results_natural_table}
\end{table}

Qualitative comparisons are shown in Figure~\ref{results_natural}. In the dehazing task, existing methods often produce images with excessive brightness after haze removal. In contrast, our method not only removes haze effectively but also preserves natural brightness levels, which yields results that are closer to the ground truth. In the deraining task, the proposed method successfully removes most of the streak-like rain while preserving more image details. Besides, the quantitative results are summarized in Table \ref{results_natural_table}. While CSUD demonstrates excellent performance on the deraining task, it struggles with other tasks, as its training paradigm is specifically designed for deraining and does not generalize well to other degradation models. As shown in Table~\ref{results_natural_table}, the proposed method consistently outperforms existing approaches. The main reason is that, rather than explicitly modeling a specific degradation type, our approach leverages classifier guidance to perform latent manifold completion.

\subsection{Performance on the Downstream Task}
Although our method achieves substantial improvements across multiple clinically relevant metrics, facilitating more accurate visual assessment for clinicians, we further evaluate its effectiveness on downstream computer-aided diagnosis tasks based on a COVID-19 CT dataset~\cite{afshar2022human}.
Specifically, we first train a ResNet18~\cite{he2016deep} on the training data and then evaluate its diagnostic performance on images enhanced using various denoising methods.
The results are reported in Table~\ref{class_results}, it can be seen compared with other image quality enhancement methods, our approach consistently achieves superior performance on downstream tasks, demonstrating its practical utility in clinical applications.

\begin{table}[t!]
  \centering
  \footnotesize
  \caption{Quantitative results of the downstream task.}
  \begin{tabular}{ccccccc}
    \toprule
     & AttentionGAN & SRC & UNSB & CSUD & Ours \\
    \midrule
    ACC~($\uparrow$) & 72.16\% & 76.13\% & 66.43\% & 61.69\% & 77.32\% \\
    \bottomrule
  \end{tabular}
  \vspace{-15pt}
  \label{class_results}
\end{table}


\subsection{Ablation Experiment and Analysis}

\begin{table}[!t]
  \centering
  \small
  \setlength{\tabcolsep}{1mm}
  \caption{Effectiveness of the proposed uncertainty-guided manifold smoothing method}
  \resizebox{\columnwidth}{!}{
  \begin{tabular}{lllllll|ll}
    \toprule
    &\multicolumn{2}{c}{LDCT} & \multicolumn{2}{c}{SVCT} & \multicolumn{2}{c}{LACT} & \multicolumn{2}{c}{Average}\\
    \cmidrule(r){2-3} \cmidrule(r){4-5} \cmidrule(r){6-7} \cmidrule(r){8-9} Algorithm & PSNR & SSIM & PSNR & SSIM & PSNR & SSIM & PSNR & SSIM\\
    \midrule
    CycleGAN & 38.25 & 94.57 & 36.01 & 91.44 & 31.97 & 90.93 & 35.41 & 92.31\\
    $\text{CycleGAN}^{\dag}$ & 39.25 & 95.30 & 36.32 & 91.95 & 32.84 & 91.76 & 36.13 & 93.00\\
    $\Delta$ & \textcolor{green!50!black}{$\uparrow$ 1.00} & \textcolor{green!50!black}{$\uparrow$ 0.73} & \textcolor{green!50!black}{$\uparrow$ 0.31} & \textcolor{green!50!black}{$\uparrow$ 0.51} & \textcolor{green!50!black}{$\uparrow$ 0.87} & \textcolor{green!50!black}{$\uparrow$ 0.83} & \textcolor{green!50!black}{$\uparrow$ 0.72} & \textcolor{green!50!black}{$\uparrow$ 0.69}\\
    \hline
    Switchable & 37.26 & 92.92 & 33.38 & 84.39 & 27.40 & 85.45 & 32.68 & 87.58\\
    $\text{Switchable}^{\dag}$ & 38.60 & 94.16 & 34.84 & 88.64 & 27.99 & 86.41 & 33.81 & 89.73\\         $\Delta$ & \textcolor{green!50!black}{$\uparrow$ 1.34} & \textcolor{green!50!black}{$\uparrow$ 1.24} & \textcolor{green!50!black}{$\uparrow$ 1.46} & \textcolor{green!50!black}{$\uparrow$ 4.25} & \textcolor{green!50!black}{$\uparrow$ 0.59} & \textcolor{green!50!black}{$\uparrow$ 0.96} & \textcolor{green!50!black}{$\uparrow$ 1.13} & \textcolor{green!50!black}{$\uparrow$ 2.15}\\
    \hline
    GcGAN & 40.33 & 96.90 & 36.14 & 92.24 & 29.36 & 89.94 & 35.27 & 93.02\\
    $\text{GcGAN}^{\dag}$ & 40.13 & 96.73 & 36.64 & 92.43 & 30.34 & 90.39 & 35.70 & 93.18\\ 
        $\Delta$ & \textcolor{gray}{$\downarrow$ 0.20} & \textcolor{gray}{$\downarrow$ 0.17} & \textcolor{green!50!black}{$\uparrow$ 0.50} & \textcolor{green!50!black}{$\uparrow$ 0.19} & \textcolor{green!50!black}{$\uparrow$ 0.98} & \textcolor{green!50!black}{$\uparrow$ 0.45} & \textcolor{green!50!black}{$\uparrow$ 0.43} & \textcolor{green!50!black}{$\uparrow$ 0.16}\\
    \hline
    SRC & 35.52 & 88.16 & 32.07 & 79.05 & 29.09 & 83.92 & 32.22 & 83.71\\
    $\text{SRC}^{\dag}$ & 35.13 & 90.34 & 33.29 & 85.77 & 30.32 & 89.78 & 32.91 & 88.63 \\ 
            $\Delta$ & \textcolor{gray}{$\downarrow$ 0.39} & \textcolor{green!50!black}{$\uparrow$ 2.18} & \textcolor{green!50!black}{$\uparrow$ 1.22} & \textcolor{green!50!black}{$\uparrow$ 6.72} & \textcolor{green!50!black}{$\uparrow$ 1.23} & \textcolor{green!50!black}{$\uparrow$ 5.86} & \textcolor{green!50!black}{$\uparrow$ 0.69} & \textcolor{green!50!black}{$\uparrow$ 4.92}\\
    \bottomrule
  \end{tabular}
  }
  \vspace{-10pt}
  \label{aug_effectiveness}
\end{table}

\noindent \textbf{{The Effectiveness of UMS.}} The UMS method is designed to construct a smooth manifold that effectively bridges discrete transitions between different sub-manifolds. To evaluate its effectiveness, the data synthesized through the UMS framework were incorporated into the training dataset, and the experimental results of different methods are summarized in Table \ref{aug_effectiveness}. The results demonstrate that UMS consistently improves the performance of multiple unsupervised image reconstruction methods across diverse domains, with only minor exceptions.

Notably, UMS provides a \textit{``free lunch"} improvement, which requires neither architectural modifications nor additional training for the base denoising models. After data generation, it can be seamlessly integrated into different denoising methods in a \textit{``free lunch"} manner, without introducing additional modules or requiring further adaptation. This plug-and-play characteristic makes it a highly versatile and efficient enhancement to existing unsupervised frameworks without introducing computational overhead or implementation complexity. 


\begin{table}[!t]
\centering
  \small
  \setlength{\tabcolsep}{1mm}
  \caption{Ablation study on joint training and contribution parts}
  \resizebox{\columnwidth}{!}{
  \begin{tabular}{lcccc|llllll}
    \toprule
    & \multicolumn{4}{c}{Dataset} &\multicolumn{2}{c}{LDCT} & \multicolumn{2}{c}{SVCT} & \multicolumn{2}{c}{LACT}\\
    \cmidrule(r){2-5} \cmidrule(r){6-7} \cmidrule(r){8-9} \cmidrule(r){10-11} 
    Algorithm & LD & SV & LA & Syn & PSNR & SSIM & PSNR & SSIM & PSNR & SSIM\\
    \midrule
    Single-LD & \ding{51} & \ding{55} & \ding{55} & \ding{55} & 38.84 & 94.95 & -- & -- & -- & --\\
    Single-SV & \ding{55} & \ding{51} & \ding{55} & \ding{55} & -- & -- & 35.35 & 90.55 & -- & --\\
    Single-LA & \ding{55} & \ding{55} & \ding{51} & \ding{55} & -- & -- & -- & -- & 31.91 & 87.92\\
    \hline
    Multi-Vanilla & \ding{51} & \ding{51} & \ding{51} & \ding{55} & 38.25 & 94.57 & 36.01 & 91.44 & 31.97 & 90.93\\
    $\text{Multi-Vanilla}^{\star}$ & \ding{51} & \ding{51} & \ding{51} & \ding{55} & 38.51 & 94.19 & 36.22 & 92.11 & 32.51 & 90.36\\
    $\text{Multi-Vanilla}^{\dag}$ & \ding{51} & \ding{51} & \ding{51} & \ding{51} & 39.25 & \textbf{95.30} & 36.32 & 91.95 & 32.84 & 91.76\\
    Mixup & \ding{51} & \ding{51} & \ding{51} & \ding{51} & 36.55 & 93.93 & 33.46 & 84.90 & 28.80 & 85.15\\
    \hline
    \rowcolor{gray!20}
    Ours & \ding{51} & \ding{51} & \ding{51} & \ding{51} & \textbf{39.66} & 95.18 & \textbf{36.89} & \textbf{92.34} & \textbf{33.79} & \textbf{92.06}\\
    \bottomrule
  \end{tabular}
  }
  \label{ablation-results}
  \vspace{-10pt}
\end{table}

\noindent \textbf{{Multi-Domain v.s. Single-Domain Training.}} To assess the comparative advantages of multi-domain versus single-domain training paradigms, a comprehensive experimental study is conducted. The corresponding results are reported in Table \ref{ablation-results}.
The first three rows correspond to training on a single NICT domain, whereas the subsequent rows correspond to joint training across all three domains.
The results of $\text{Multi-Vanilla}$ reveal that joint training improves quantitative performance across multiple domains, with particularly notable gains in SVCT and LACT images.

$\text{Multi-Vanilla}^{\star}$ refers to a variant in which the generator is replaced with the proposed confidence-guided global- and sub-manifold-driven architecture. Meanwhile, $\text{Multi-Vanilla}^{\dag}$ denotes the application of UMS to smooth the manifold. Besides, the results demonstrate that both components contribute significantly to performance gains over the baseline. Notably, $\text{Multi-Vanilla}^{\dag}$ yields more pronounced improvements, which aligns with the earlier hypothesis. This observation further supports the claim that the primary challenge lies in modeling discrete manifolds. Once the manifold is smoothed via our proposed approach, reconstruction performance can be substantially enhanced. 
Furthermore, Mixup augmentation is applied to expand the dataset by linearly interpolating training samples and their labels across different classes.
However, due to the varying noise distributions induced by different scanning protocols, Mixup fails to deliver satisfactory performance, highlighting the limitations of naive sample interpolation under protocol-induced domain shifts. Our method integrates the data generation-based UMS with the proposed denoising architecture, and ablation studies demonstrate that each component contributes meaningfully to the overall performance, resulting in superior results across various evaluation metrics.

\begin{figure}[t!]
  \centering
  \includegraphics[width=0.9\linewidth]{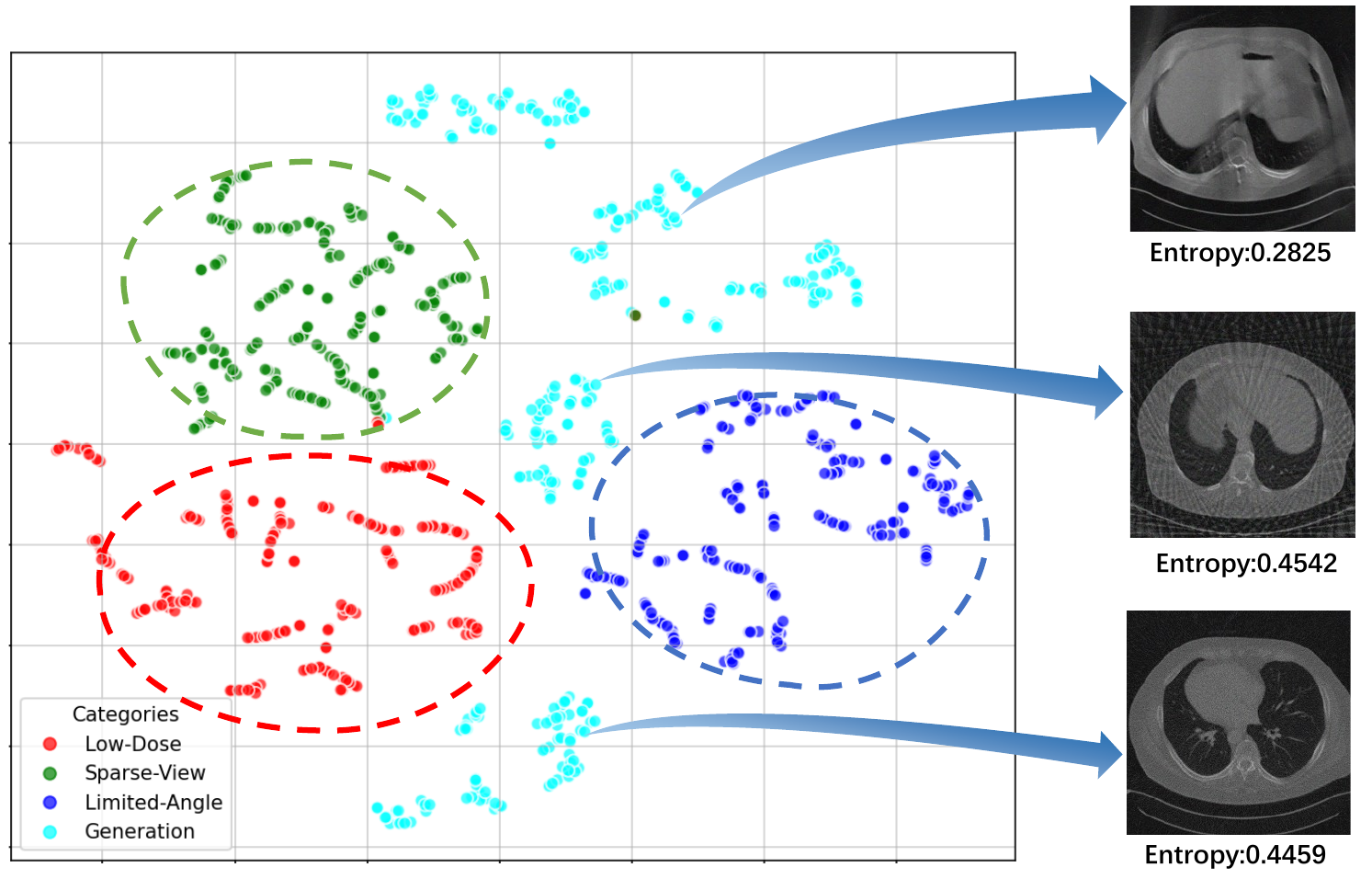}
  \caption{t-SNE visualization of real and synthetic data.}
  \label{t-sne}
  \vspace{-10pt}
\end{figure}

\noindent \textbf{{Assumption Clarification.}}
As our method is classifier-guided and the classifier is optimized using the cross-entropy loss, training encourages confident predictions by maximizing the likelihood of the correct class. As a result, well-classified samples are driven toward peaked predictive distributions with low entropy near class centers. In contrast, samples near decision boundaries yield higher loss, flatter predictions, and consequently higher entropy. To empirically validate this behavior, we present t-SNE visualizations of both real and synthetic data in Figure~\ref{t-sne}. Different scanning protocols form distinct sub-manifolds, with high-entropy samples concentrated near the manifold boundaries. By augmenting the training set, these protocol-specific sub-manifolds are effectively bridged, yielding a smoother and connected global manifold.

\section{Conclusion}
\label{sec:con}
In this paper, we propose a  CT unified model training framework to rule all scanning protocols, which addresses the fundamental challenge of manifold discontinuity across different NICT acquisition strategies. Firstly, we treat the discrete heterogeneous sub-manifold as components of a complex manifold, and design UMS to smooth the manifold. Besides, we design a novel global- and sub-manifold-driven architecture to restore higher quality images through modeling the global- and sub-manifold information simultaneously. Extensive experiments demonstrate our effectiveness. Besides, the smoothing manifold operation can consistently improve the performance for different unsupervised methods without requiring architectural modifications. In future work, extending this methodology to other medical imaging modalities with distinct physical principles represents an interesting research field.


\bibliographystyle{ACM-Reference-Format}
\bibliography{main-sample-sigconf-authordraft}

\appendix









\end{document}